\UseRawInputEncoding
\documentclass{article} 
\usepackage[preprint]{colm2026_conference}

\usepackage{tikz}
\usetikzlibrary{arrows.meta,positioning,fit,calc}

\usepackage{microtype}
\usepackage{hyperref}
\usepackage{url}
\usepackage{booktabs}

\usepackage{graphicx}
\usepackage{amssymb}

\usepackage{lineno}
\usepackage{placeins}
\usepackage{caption}
\usepackage{ulem}
\usepackage{threeparttable}
\usepackage{pifont}
\usepackage[table]{xcolor}
\usepackage{colortbl}
\newcommand{\heat}[1]{%
  \cellcolor{purple!#1!white!70}#1%
}

\usepackage{wrapfig} 
\usepackage[most]{tcolorbox}
\definecolor{PromptTitle}{HTML}{3973B3} 
\tcbset{
  colback=white,
  colbacktitle=PromptTitle,
  colframe=black!30,
  boxrule=0.6pt,
  arc=2mm,
  left=1.2mm,right=1.2mm,top=1mm,bottom=1mm,
  fonttitle=\bfseries,
}
\usepackage{fvextra}
\usepackage{needspace}
\definecolor{darkblue}{rgb}{0, 0, 0.5}
\hypersetup{colorlinks=true, citecolor=darkblue, linkcolor=darkblue, urlcolor=darkblue}

\title{Reasoners or Translators? Contamination-aware Evaluation and Neuro-Symbolic Robustness in Tax Law}

\author{
Parisa Kordjamshidi$^{1,2}$\thanks{This work was done during a sabbatical at Bloomberg} \\
\And
Samer Aslan$^{1}$ 
\And
Madhavan Seshadri$^{1}$ 
\And
Leslie Barrett$^{1}$ 
\And
Enrico Santus$^{1}$ 
\\
$^{1}$Bloomberg \texttt{\{pkordjamshid,saslan10,mseshadri,lbarrett4,esantus\}@bloomberg.net}\\
$^{2}$Michigan State University
}

\begin{document}

\ifcolmsubmission
\linenumbers
\fi

\maketitle


\begin{abstract}
Recent advances in large language models (LLMs) have significantly enhanced automated legal reasoning. Yet, it remains unclear whether their performance reflects genuine legal reasoning ability or artifacts of data contamination. We present a comprehensive empirical study of tax law reasoning approaches and implement a contamination detection protocol to rigorously assess LLM reliability. We show that performance can be inflated by contamination. 
Building on this analysis, we conduct a systematic evaluation, comparing monolithic LLMs with hybrid systems that translate statutory text into formal representations and delegate inference to symbolic solvers. We build a novel test suite designed to probe generalization to unseen documents via case and rule variations. 
Our findings indicate that legal reasoning is inherently compositional and that neuro-symbolic frameworks offer a more reliable and robust foundation for legal AI, as well as improved generalization to unobserved situations.
\end{abstract}

\section{Introduction}

Legal reasoning is a complex and sensitive task that requires rigorous, explainable, and verifiable frameworks, as errors can have severe and irreversible consequences. For example, mistakes in tax returns may affect credit, student aid eligibility, and health coverage. While recent large language models (LLMs) have shown impressive capabilities~\citep{brown2020language,openai2023gpt4}, their performance on legal reasoning tasks remains inconsistent~\citep{chalkidis2020legalbert,10.5555/3666122.3668037} and prone to hallucination~\citep{10.1093/jla/laae003,magesh2025hallucination}.
Reported accuracies vary widely across benchmarks, with some nearing 90\%~\citep{10.1098/rsta.2023.0159} and others falling below practical thresholds~\citep{NGUYEN2025106165}. More broadly, inflated performance in LLM evaluations is often attributed to data contamination and memorization~\citep{10.1162/TACL.a.20}, raising concerns about benchmark validity and undermin1ing reliability in genuinely novel legal scenarios, where generalization to unseen statutes, fact patterns, and interpretations is essential.


\begin{figure*}
\centering
  \includegraphics[width=0.9\textwidth]{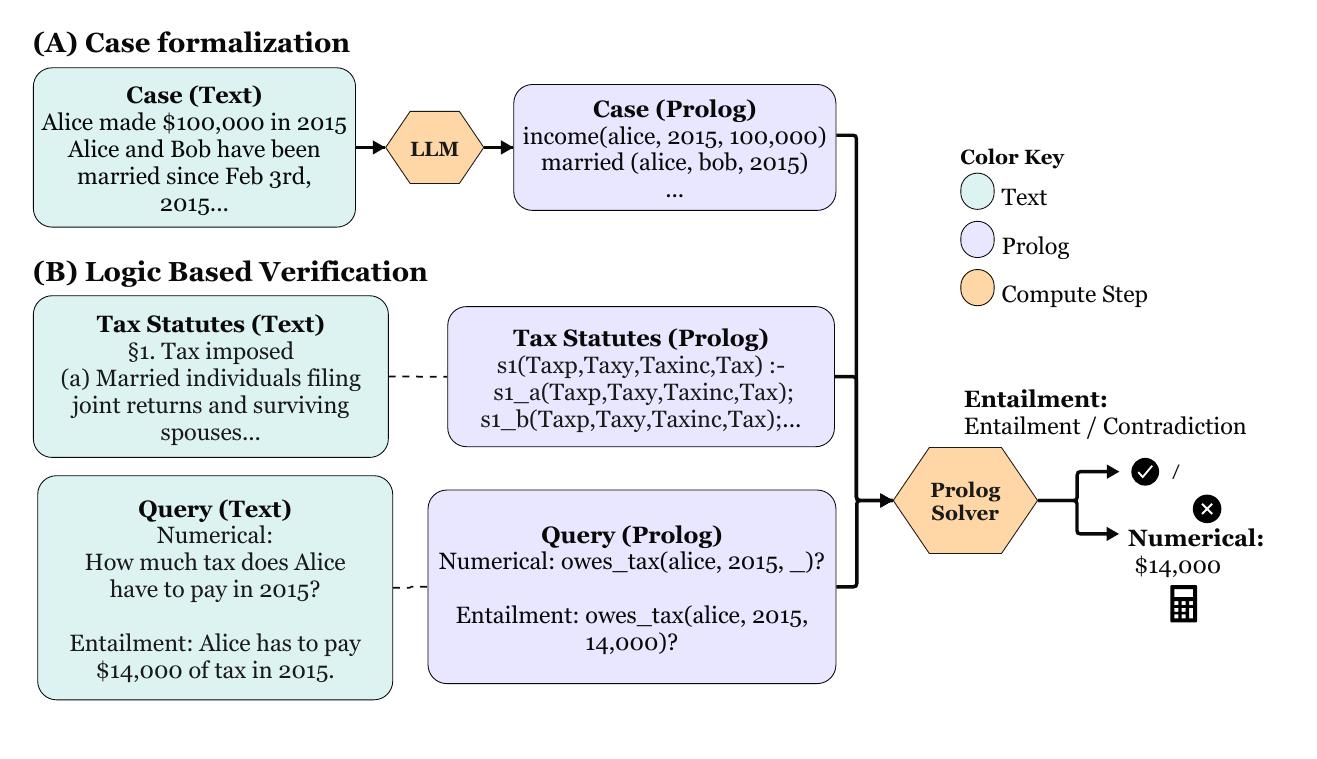}
  \caption{\textbf{Neuro-symbolic pipeline for tax-law reasoning.}
(A) \emph{Case formalization:} an LLM converts a textual fact pattern into a set of Prolog facts (structured case knowledge).
(B) \emph{Logical reasoning:} Given the Prolog representation of the statute excerpt and a query, the Prolog solver executes the composed program (case facts + statute rules + query) to produce either a Boolean decision (entailment vs.\ contradiction) or a numeric answer (e.g., tax owed).} 
 \label{fig:teaser}
\end{figure*}

Recent work in LLM-based legal reasoning has focused on making reasoning explicit through neuro-symbolic models to improve verification and interpretability. These approaches combine symbolic reasoning, which is explainable and verifiable, with neural models capable of handling complex, real-world data. In tax law, this is typically achieved by coupling LLMs with symbolic engines such as Prolog, SAT solvers~\citep{Holzenberger2020Dataset,feng2025vericotneurosymbolicchainofthoughtvalidation}, and domain-specific languages~\citep{lorenzo-etal-2025-translating}. More broadly, the integration of general solvers and theorem provers has been proposed to further enhance formal reasoning capabilities of LLMs~\citep{jiang-etal-2024-leanreasoner,pan-etal-2023-logic}.
A central bottleneck, however, is translating natural language into formal logic. While such formalization has long been viewed as a principled foundation for verifiability and accountability in legal analytics~\citep{Ashley_2017}, earlier studies have shown that scaling logical reasoning to real-world applications is challenging~\citep{Sartor2010}. More recent work in the LLM era has further highlighted the difficulty of mapping natural language to formal logic~\citep{yang-etal-2024-harnessing,putra2026nl2logic}. Although advances in semantic parsing with LLMs have improved legal text annotation~\citep{Savelka_2023}, empirical results in tax law show that full formal translation remains challenging. In some case studies, pipelines relying on explicit formalization can be outperformed by direct natural language reasoning~\citep{jurayj2025languagemodelslogicprograms}, highlighting limitations of current approaches in realistic settings. 
Despite these limitations, a key advantage of this line of work is enabling verification and effective human-in-the-loop intervention. Prior work shows that validating generated logical programs and deferring uncertain cases to experts can substantially reduce tax penalties~\citep{jurayj2025languagemodelslogicprograms}, underscoring the value of explicit reasoning even when full automation is not feasible. Agentic frameworks further support feedback, verification, and self-correction, improving the robustness of hybrid neuro-symbolic systems~\citep{hitzler2022neuro_symbolic}.

In this paper, we present a systematic study of legal reasoning in tax law. We revisit existing approaches and reproduce prior results using state-of-the-art LLMs. Motivated by the risks of data contamination, we compare the reasoning abilities of monolithic LLMs with neuro-symbolic models and revisit these approaches. 
Specifically, we adopt a modular approach that involves translating textual cases into logical facts, maintaining a knowledge base (KB) of logical rules, and forming logical programs by combining these facts and rules. 
Figure~\ref{fig:teaser} illustrates the query \textit{"How much tax should Alice pay in 2015?"}, translated to $owes_tax(alice, 2015,_)$ in Prolog. Executing this query over the formalized statutes and case resolves the unknown argument, yielding a tax value of $14{,}000$. In this setup, LLMs serve as translation tools and deductive reasoning is handled by Prolog.

The goal of our extensive empirical evaluation is to address:\textit{ \textbf{(Q1)} the effectiveness of monolithic vs.\ neuro-symbolic approaches, \textbf{(Q2)} alignment with prior results and the performance of newer state-of-the-art LLMs, \textbf{(Q3)} the impact of data contamination, and \textbf{(Q4)} generalization under case and rule perturbations.} 
Our findings show that LLM performance is partly inflated by contamination and lacks stability under variations. In contrast, neuro-symbolic models are more robust and reliable, achieving better performance under rule and data changes. We conclude that legal reasoning remains a compositional and complex task and advocate for hybrid neuro-symbolic systems that combine LLM capabilities in parsing and formal translation with symbolic reasoning engines to improve generalization, interpretability, and verifiability.

\section{Related Work}
\label{sec:Related-work}

Legal reasoning has become an active area of research with the rise of LLMs. It provides a rich test bed for diverse reasoning types, including case-based (analogical), abductive, and deductive reasoning~\citep{NGUYEN2025106165}, often used in combination. Consequently, numerous benchmarks, datasets, and model architectures have been proposed to evaluate legal reasoning systems~\citep{chalkidis-etal-2022-lexglue,10.5555/3666122.3668037,shi2026plawbenchrubricbasedbenchmarkevaluating}. 
Here, we focus on prior work in tax law. We choose this domain due to the availability of human annotations and its structured, rule-based nature, making it well suited for neuro-symbolic modeling. 
In tax law, the most widely used dataset is $SARA$~\citep{Holzenberger2020Dataset}, which provides human-annotated logical forms of rules and cases. Recent LLM performance has been evaluated by~\citet{hu-etal-2025-evaluating} across legal tasks in Chinese and English, including tax law reasoning on $SARA$, highlighting challenges such as hallucinations and limited legal knowledge.

The use of symbolic solvers in law dates back to early AI research~\citep{buchanan1970ai_legal_reasoning}, with logic programming and rule-based systems widely applied to legal reasoning~\citep{sergot1986british_nationality,schild1990open}. Much of this work emerged from argumentation mining and computational argumentation research~\citep{10.1007/s10506-010-9104-x,COLLENETTE2023103861}. More recent research focuses on translating legal text into formal logic, using LLMs to obtain executable representations. 
In this direction,~\citet{jurayj2025languagemodelslogicprograms} propose a Prolog-based framework, arguing that verifiable programs with human-in-the-loop supervision can mitigate monetary losses from tax computation errors.

With the rise of agentic AI, integrating formal logic into agentic frameworks has been explored~\citep{10.1145/3746252.3761057}. In this approach, cases are translated into formal logic, and an SMT solver computes numerical tax values; however, evaluation is limited to the numerical subtask, with no entailment results reported. Beyond statutory reasoning, case-based analogical methods have also been studied in tax law, relying on comparisons with prior cases rather than explicit rules. For example,~\citet{Zou2024ReframingTL} propose a pairwise comparison framework using textual similarity for entailment, though their results are less competitive than recent LLM-based approaches.
%
%
\section{Evaluation Framework} 

Recent work on legal reasoning with large language models highlights two key challenges affecting both fine-tuned and test-time approaches~\citep{10.1093/jla/laae003}. First, models often rely on outdated or inaccurate knowledge due to memorization of contaminated data. Second, they tend to hallucinate facts or legal conditions not grounded in the input. Beyond these issues, reliability and verifiability remain critical for real-world deployment, where errors can have serious consequences. 
These limitations motivate a neuro-symbolic framework, which combines the rigor of symbolic reasoning with the flexibility of LLMs: solvers perform verifiable reasoning, while LLMs handle parsing and formalization of natural language. The pipeline in Figure~\ref{fig:teaser} implements this approach and enables systematic comparison with standalone LLM methods.  
We use an LLM to convert the natural language descriptions of legal cases into factual assertions in predicate logic, instantiated in Prolog~\citep{10.1145/234286.1057820}. The question is similarly translated into a logical query, which may be fully instantiated or contain variables whose bindings form the answer. Statutes, originally in natural language, define a textual knowledge base of legal rules. We use a human-coded Prolog Statute KB provided with SARA. The resulting rules and facts are then executed by the logical reasoning engine.

\subsection{Data Contamination} 

\label{sec:contamination}

We hypothesize that the strong performance of recent LLMs in the legal domain is partly due to data contamination and benchmark memorization. Much of the publicly available legal data may have been incorporated—directly or indirectly—into training corpora, compromising evaluation validity and motivating contamination-free test sets. 
To investigate this, we conduct data contamination analyses using adapted detection techniques~\citep{10.1162/TACL.a.20}. We also construct a new synthetic test set based on $SARA$ to minimize contamination while preserving legal and structural complexity. Given the high cost of human annotation in legal reasoning, this dataset is a key contribution. We describe both the analysis and data generation process below.

\subsubsection{Data Contamination Test}

We follow~\citep{10.1162/TACL.a.20} to construct contamination test sets in a multiple-choice format, where models must select the option matching the $SARA$ dataset. Two sets are created: the Bias Detector Quiz (BDQ) and the Bias Compensator Quiz (BCQ). In BDQ, each question includes four perturbed versions of a $SARA$ instance—generated via word-level LLM perturbations that preserve meaning—along with a fifth "none of the above" option. This setup is used to identify positional biases. Based on these results, BCQ is constructed by placing correct answers in "non-preferred" positions that the model selects with probability lower than random chance. 

In BCQ, one of the four perturbations is replaced with the original $SARA$ instance. This process is repeated for each non-preferred position identified in BDQ to create multiple test splits. Model performance on these splits is used to estimate contamination while accounting for positional bias. To further mitigate this bias, following \cite{10.1162/TACL.a.20}, we apply chance-adjusted accuracy using Cohen’s Kappa, $\kappa = \frac{p_o - p_e}{1 - p_e}$, where $p_o$ is the observed agreement, i.e., the best BCQ performance, and $p_e$ is the expected agreement, i.e. the probability of selecting the non-preferred option in BDQ where the LLM achieved its best performance by placing the correct answer in that position. We use $\kappa$ as the contamination measure. 
For a more conservative estimate, we also report minimum and maximum contamination levels. The minimum corresponds to the lowest $\kappa$ across non-preferred positions, the most conservative estimate, and the maximum to the highest $\kappa$ across BCQ splits, reflecting the strongest contamination signal.

\subsection{Generating New Test Data: $SARA^+$}

\label{sec:SARA+} 
Our primary dataset for tax law reasoning is $SARA$~\citep{Holzenberger2020Dataset}. Its structured, rule-based nature reduces ambiguity and enables clearer analysis of logical rule application. Additionally, it includes costly, human-coded logical programs and rules, offering strong potential for expansion. 
We develop multiple new versions of $SARA$ with two objectives: mitigating data contamination by introducing novel tax reasoning challenges, and enabling synthetic data generation with high-quality formal annotations. This augmentation supports both rigorous evaluation and future training. We describe the original $SARA$ dataset and our approach for generating the splits categorized in Table~\ref{tab:sara+}. 

\begin{table}
\small
\centering
\renewcommand{\arraystretch}{1.3}
\begin{tabular}{|lccc|}
\hline
\textbf{Task} & \textbf{Train} & \textbf{Test} & \textbf{Total} \\
\hline
Entailment & 176 & 100 & 276 \\
Numeric    & 80  & 20  & 100 \\
\hline
\end{tabular}
\caption{Dataset statistics for the $SARA$ (StAtutory Reasoning Assessment).}
\label{tab:sara_stats}

\end{table}

\noindent\textbf{$SARA$:} The $SARA$ dataset consists of statutory provisions in natural language with human-annotated formal representations as Prolog Horn clauses, denoted as $r$. Each instance includes (i) a natural language case description and (ii) its formal representation as a Prolog KB of facts, denoted as $c$. 
Two tasks are defined over this dataset. One is textual entailment ($SARA_E$), which determines whether a statement is entailed by the case under the statutes, and the other is numerical reasoning ($SARA_N$), which computes the tax liability implied by the case. In both tasks, the query is manually formalized as a Prolog query $q$, and the answer is either a binary decision or a numerical value. Illustrative examples are shown in Figure~\ref{fig:teaser}. 

\noindent\textbf{Rule Perturbation}: We create a new set of Prolog rules, denoted $r^\prime$, by conservatively perturbing numerical values in both the Prolog statutes and their corresponding textual rules using an LLM, with manual verification to ensure consistency across modalities. 

\noindent\textbf{Case Perturbation}: Based on both $r$ and $r^\prime$, we generate new cases, queries, and answers using the entailment and numerical splits of $SARA$. The entailment split consists of two subsets: a subset of 100 cases requiring numerical reasoning ($SARA_{En}$) and 172 cases requiring only conceptual reasoning~\footnote{There are four additional examples, held out for in-context use.}, e.g., \textit{dependent} ($SARA_{Ee}$).

\noindent {$\mathbf{SARA^r_E}$:}
In this split, we use the original entailment cases ($SARA_E$) and their textual forms, but generate answers by executing the corresponding Prolog programs under the modified rule set $r^\prime$. As a result, the underlying reasoning and tax computations change, leading to different answers.

\noindent {$\mathbf{SARA^r_N}$: } The cases are the same as the original $SARA_N$ split, while the rule set is replaced with $r^\prime$. The procedure for constructing this test split follows that of $SARA^r_E$. 

\noindent {$\mathbf{SARA^c_{N}}$:} 
We use cases from the $SARA_{N}$ split and perturb them using LLMs by modifying numerical values in both the text and corresponding Prolog programs within a range of $\pm 30\%$. The original rule set $r$ is retained. These changes can alter applicable rules and resulting tax amounts. 
We recompute taxes using the updated Prolog programs under $r$ to obtain positive entailment examples. Negative examples are constructed by introducing a $\delta$ offset of approximately $\pm (1\text{--}5)\%$ of the correct tax amount and using this perturbed value to form incorrect queries.

\noindent{$\mathbf{SARA^c_{En}}$:} 
We apply the same procedure used for constructing $SARA^c_{N}$ to the $SARA_{En}$ subset, resulting in $SARA^c_{En}$.

\noindent {$\mathbf{SARA^c_{Ee}}$:} 
We use cases from the $SARA_{Ee}$ split and perturb them via LLM-based paraphrasing while preserving semantics and numerical values. The original rule set $r$ is retained, so the Prolog programs remain unchanged. Consequently, queries and answers are identical to the original ones, allowing reuse of the same positive and negative examples with paraphrased text.

\noindent{$\mathbf{SARA^{rc}_{N}}$:} 
For this split, we use the same cases as in $SARA^c_{N}$, with modified numerical values, but apply the updated rule set $r'$. As the rules differ, the resulting tax values change compared to both $SARA^c_{N}$ and the original $SARA$. The taxes are computed using the updated Prolog programs under $r'$, yielding positive entailment examples.
Negative examples are constructed as in $SARA^c_{N}$ by adding a $\delta$ offset of approximately $\pm (1\text{--}5)\%$ to the correct tax and forming incorrect queries.

\noindent {$\mathbf{SARA^{rc}_{En}}$:}
For this split, we apply the same procedure as for $SARA^{rc}_{N}$, restricted to the $SARA^c_{En}$ subset of the entailment split.

\begin{table}

\centering
\renewcommand{\arraystretch}{1.2}
\small
\begin{tabular}{|l|c|c|c|c|}
\hline
 & \textbf{Orig} & \textbf{RC} & \textbf{CC} & \textbf{CP} \\
\hline
\multicolumn{5}{|l|}{\textbf{Numerical}} \\
\hline
$SARA_{N}$        &\textbf{X} &   &   &   \\
\hline
$SARA_{N}^{r}$    &   & \textbf{X}&   &   \\
\hline
$SARA_{N}^{c}$    &   &   & \textbf{X} &   \\
\hline
$SARA_{N}^{rc}$   &   & \textbf{X} & \textbf{X} &   \\
\hline
\multicolumn{5}{|l|}{\textbf{Entailment} ($e$: no numerical reasoning)} \\
\hline
$SARA_{Ee}$       & \textbf{X} &   &   &   \\
\hline
$SARA_{Ee}^{c}$   &   &   &   & \textbf{X} \\
\hline
\multicolumn{5}{|l|}{\textbf{Entailment} ($n$: numerical reasoning)} \\
\hline
$SARA_{En}$       &\textbf{X}&   &   &   \\
\hline
$SARA_{En}^{r}$   &   & \textbf{X} &   &   \\
\hline
$SARA_{En}^{c}$   &   &   & \textbf{X} &   \\
\hline
$SARA_{En}^{rc}$  &   & \textbf{X} & \textbf{X} &   \\
\hline
\end{tabular}

\caption{Novel test splits in $SARA^+$. 
Orig = Original Dataset; 
RC = Rule Change; 
CC = Case Numerical Change; 
CP = Case Paraphrasing}
\label{tab:sara+}
\end{table}

In all cases, identical perturbations are applied to both textual rules and Prolog programs, ensuring alignment between the formal problem and its solutions. This enables precise formal representations and exact solutions without human expert effort. To maintain consistency, we restrict modifications to numerical values only. Although minimal and controlled, these changes can substantially affect reasoning, as rule applicability often depends on quantities such as salaries and years. We hypothesize that this approach yields challenging test cases for LLMs.

\section{Experimental Results}

\textbf{Datasets.} 
We use the \textbf{$SARA$} dataset~\citep{Holzenberger2020Dataset}, summarized in Table~\ref{tab:sara_stats}, and \textbf{$SARA^{+}$}, our novel test introduced in Section~\ref{sec:SARA+}.\footnote{The SARA$^+$ dataset is publicly available at \url{https://github.com/HLR/Legal-Reasoning}.}\\

\noindent\textbf{Evaluation Metrics.}
For entailment, we use accuracy ($Acc$), defined as the proportion of correct binary decisions. For the numerical task, we use Exact Match ($EM$), which requires the predicted numerical value to match the ground truth. Some related work reports a relaxed metric ($M10\%$), counting predictions within a 10\% window of the ground truth as correct. 
In prior solver-based work, models may abstain via code verification; in such cases, both $EM$ and $Err$ are reported, where $Err$ counts incorrect non-abstained predictions.\\
\noindent \textbf{LLMs for Data Generation.} 
We use Claude Sonnet 4.5 for creating the contamination test splits and GPT-5.2 to generate $SARA^+$. The new rules and a subset of new cases are manually verified for quality control. For task-solving experiments, we evaluate a range of LLMs, including both proprietary and open-source models, as reported in the results tables.

\subsection{Baselines from Previous Research}

We investigate the following question: \noindent\textbf{Q1. Which approach is more effective for legal reasoning: monolithic LLMs or neuro-symbolic models?} 
We select representative prior work and metrics to summarize the state of the art across diverse approaches and backbone LLMs, including both reasoning and non-reasoning models. As shown in Table~\ref{tab:SOTA-direct}, reasoning models such as DeepSeek-R1 achieve state-of-the-art performance, reaching 91.79\% accuracy on entailment. For numerical reasoning, results are reported using mean squared error, which is not directly comparable to other metrics; nevertheless, DeepSeek-R1 performs best, followed by o1-preview. 
\begin{table}

\centering
\small
\begin{tabular}{|lcc|}
    \hline
    Model & $SARA_E$  & $SARA_N$ \\
    \hline
    GPT-4o & 87.87 & 1.2073 \\
    o1-preview & 91.18 & 1.0930 \\
    o1-mini & 89.34 & 1.3762 \\
    DeepSeek-V3 & 83.09 & 2.3079 \\
    DeepSeek-R1 & 91.79 & 0.2460 \\
    Llama3.1-405B & 80.88 & 7.7170 \\
    Qwen2-72B-Instruct & 85.29 & 4.8066 \\
    QwQ-32B-Preview & 71.32 & 3.3034 \\
    Claude Sonnet 4 & 89.30 & 6.95 \\
    GLM-zero-preview & 90.77 & 7.7855 \\
    \hline
\end{tabular}

\caption{Direct QA results on $SARA$ selected from~\citet{hu-etal-2025-evaluating}, SARA$_N$ in MSE.}
\label{tab:SOTA-direct}

\end{table}
\begin{table}[t]
\begin{minipage}[t]{0.40\textwidth}
\small
\begin{tabular}{|lc|}
\hline
Model & $SARA_N$ (M10\%) \\
\hline
GPT-4.1 & 79.2 \\
DeepSeek-V3 & 82.30 \\
o4-mini & 87.50 \\
Qwen 2.5 (72B) & 78.10 \\
\hline
\end{tabular}
\captionof{table}{SOLAR~\citet{10.1145/3746252.3761057}}
\label{tab:solar}

\end{minipage}
\hfill
\begin{minipage}[t]{0.55\textwidth}
\small
\begin{tabular}{|lccc|}
\hline
Model & Direct & Prolog & Prolog$^*$ \\
\hline
GPT-5 & 76/24 & 53/13 & 50/2 \\
GPT-4.1 & 48/52 & 39/31 & 38/1 \\
o3 & 56/44& 75/15& 68/7 \\
DeepSeek-V3 & 22/78 & 11/43 & 9/1 \\
DeepSeek R1 & 74/26 & 38/10 & 19/2 \\
\hline
\end{tabular}

\captionof{table}{Prolog-based $SARA_N$ (EM/Err),\\ $prolog^*$ uses ground-truth rules ~\citet{jurayj2025languagemodelslogicprograms}}
\label{tab:prolog}

\end{minipage}
\end{table}
\textbf{SOLAR} (Table~\ref{tab:solar}) is an agentic framework that constructs logical representations with LLMs and uses external SAT solvers for reasoning. Results are reported only for the numerical task using $M10\%$, a lenient metric not directly comparable to exact-match used in other work. 

\begin{table}

\centering
\small
\begin{tabular}{|lcccc|}
\hline
Model 
& \textbf{Dir-E} 
& \textbf{Pro-E} 
& \textbf{Dir-N}
& \textbf{Pro-N} \\
\hline
Gemini 3 Pro         & \textbf{94.85} & 86.80 & 81.25 & 87.5\\
o4-mini              & 93.75 & 82.00 & 59.38 & 85.4 \\
GPT-5.1 (M)     & 93.75 & 85.70 & \underline{\textbf{82.29}} & 83.33 \\
C.-sonnet-4.5    & 93.01 & 84.90 & 69.79 & 83.30\\
o1                   & 92.65 & 84.20 & 71.88 & \underline{79.2} \\
GPT-5                & 92.65 & 86.40 & 80.21 & 83.33 \\
Gemini-2.5-pro       & 90.44 & 86.03 & 72.92 & 81.30 \\
GPT-5.2 (M)     & \underline{89.71} & 85.70 & 72.92 & \underline{\textbf{87.5}} \\
C.-sonnet-4      & 89.34 & 83.80 & 55.21 & 85.40 \\
GPT-4.1              & 85.66 & 83.50 & 41.67 & 81.30 \\
GPT-4o               & 83.46 & 82.00 & \underline{10.42} & 82.30 \\
Llama 3.1 405b       & 83.7 & 81.1 & 8.3 & 71.9 \\
\hline
\end{tabular}

\caption{Comparing Direct QA (Dir) to Prolog (Pro)}
\label{tab:llms_direct_vs_prolog}

\end{table}
Finally, the \textbf{Prolog-based} results~\cite{jurayj2025languagemodelslogicprograms} (Table~\ref{tab:prolog}) do not improve over the direct baseline in correct decisions. However, they substantially reduce errors by allowing abstention when generated code fails verification. Overall, prior work suggests that standalone, reasoning LLMs achieve the highest accuracy. While agentic and symbolic frameworks offer advantages in interpretability, verification, and reliability, they do not clearly outperform strong LLMs on $SARA$. Due to differences in LLM backbones and evaluation metrics, we report in-house baselines in the next section for a more consistent comparison.

\subsection{Baselines with In-house Execution} 
We address the following question: \textbf{Q2. Do our in-house results align with prior findings, and how do newer LLMs perform?}\\
\noindent\textbf{Direct QA.} We evaluate most models from Tables~\ref{tab:solar}–\ref{tab:SOTA-direct} using direct question answering with all statutes provided in context. We reproduce prior results and extend the analysis to newer LLMs, including OpenAI (GPT-x and O-x), Claude, and Gemini models. As shown in Table~\ref{tab:llms_direct_vs_prolog}, entailment is consistently easier than numerical reasoning in Direct QA. Newer models perform better on both tasks, suggesting that scale remains a key factor for reasoning. Overall, reasoning models outperform general LLMs, with \textit{Gemini 3 Pro} leading on entailment and \textit{GPT-5.1} on numerical reasoning. Performance on the numerical task varies widely across models, from 10.42 (GPT-4o) to 82.29 (GPT-5.1), whereas entailment shows much less variation. Even recent models such as GPT-5.2 achieve only around 73\% on numerical reasoning, indicating its continued difficulty. Overall, these results suggest that entailment is less demanding and that recent models may be overfitting to this setting.

\noindent \textbf{Mapping to Logical Solvers.}
For these baselines, we follow~\cite{jurayj2025languagemodelslogicprograms} and use LLMs to translate case descriptions into Prolog predicate logic. Human-coded rules are then combined with the translated programs and executed by a Prolog solver to perform entailment or compute tax values. As shown in Table~\ref{tab:llms_direct_vs_prolog}, Prolog-based reasoning significantly outperforms Direct QA on numerical inference when using the same LLMs as translators. In contrast, Direct QA remains strong on entailment and is harder to surpass. This suggests that LLMs are more reliable as translators into formal logic than as standalone reasoners, especially as task complexity increases. 
The largest gains are observed for GPT-4o, which performs poorly under Direct QA for numerical reasoning but improves by approximately 72\% when used as a translator into Prolog. Similarly, GPT-4.1 and Claude Sonnet 4 improve by roughly 30-40\%. More advanced reasoning models, such as \textit{GPT-5.1} and \textit{Gemini 3 Pro}, also benefit from Prolog-based reasoning, though gains are smaller (1-5\%). 

Overall, results in Table~\ref{tab:llms_direct_vs_prolog} align with prior findings (Tables~\ref{tab:solar}–\ref{tab:SOTA-direct}), confirming that scaling and reasoning optimization improve performance. However, our experiments show that symbolic reasoning remains essential for robust numerical inference. 
In contrast, entailment performance appears to be saturating, potentially due to contamination. We therefore conduct a systematic analysis to quantify contamination and assess its impact across settings.

\begin{table}

\centering
\small
\begin{tabular}{|lcc|}
    \hline
    Model & Min Cont & Max Cont \\
    \hline
    \multicolumn{3}{|l|}{\textbf{Reasoning-Optimized Models}} \\
    \hline
    Gemini 3 Pro & 90.5\% & 92.0\% \\
    GPT-5.2 medium & 67.1\% & 73.0\% \\
    GPT-5.2 low & 63.7\% & 67.0\% \\
    Claude Opus 4 & 60.2\% & 63.0\% \\
    o3 & 62.4\% & 68.0\% \\
    o4-mini & 56.6\% & 64.0\% \\
    o3-mini & 34.4\% & 41.0\% \\
    \hline
    \multicolumn{3}{|l|}{\textbf{General / Balanced LLMs}} \\
    \hline
    Gemini 2.5 Pro & 77.8\% & 82.0\% \\
    Claude Sonnet 4.5 & 55.7\% & 61.0\% \\
    GPT-5.1 & 46.9\% & 57.0\% \\
    Claude Sonnet 4 & 34.9\% & 44.0\% \\
    GPT-4.1 & 30.8\% & 37.0\% \\
    GPT-4o & 24.5\% & 29.0\% \\
    \hline
\end{tabular}

\caption{Data contamination (min/max contamination) using MCQ variations of $SARA$.}
\label{tab:contam_metrics}

\end{table}

\subsection{Data Contamination Analysis}

In this section, we evaluate the data contamination hypothesis by addressing \textbf{Q3: Is the observed performance of LLMs inflated by contamination?}
We use the MCQ variant of $SARA$ (Section~\ref{sec:contamination}) to evaluate contamination. Table~\ref{tab:contam_metrics} reports $Min$ and $Max$ contamination. Results show substantial variation across models, with newer frontier models (e.g., Gemini 3 Pro, GPT-5.2, Claude Opus 4) exhibiting higher contamination than earlier models (e.g., Gemini 2.5 Pro, GPT-4.1, Claude Sonnet 4). This trend may reflect greater exposure to $SARA$-like data in recent web-scale training corpora. However, it does not imply a causal link to model scale, as models differ in architecture, training data, and post-training, and parameter counts are often unavailable. Considering these results alongside Table~\ref{tab:llms_direct_vs_prolog}, a clearer pattern emerges: contamination is strongly associated with performance in the Direct QA setting for the entailment task. 
For example, Gemini 3 Pro exhibits high contamination (approximately 90\%) and achieves the highest Direct QA performance (94.85\%), while models with lower contamination, such as GPT-4.1 (approximately 30\%), achieve lower entailment accuracy (85.66\%). A similar pattern is observed for numerical reasoning: Gemini 3 Pro achieves 81.25\%, whereas GPT-4.1 attains only 41\%, although the correlation appears weaker overall.

\begin{table*}[h]
\centering
{\fontsize{6.2}{12}\selectfont
\bfseries
\begin{tabular}{lcccccccccc}
\hline
Model
& $SARA_{En}$
& \textbf{$SARA^c_{En}$}
& \textbf{$SARA^r_{En}$}
& \textbf{$SARA^{rc}_{En}$}
& SARA$_{N}$
& \textbf{SARA$^c_{N}$}
& \textbf{SARA$^r_{N}$}
& \textbf{SARA$^{rc}_{N}$}
& SARA$_{Ee}$
& \textbf{SARA$^c_{Ee}$}\\
\hline
GPT-5
& \heat{98} & \heat{85} & \heat{85} & \heat{73}
& \heat{80} & \heat{76} & \heat{77} & \heat{74}
& \heat{90} & \heat{90} \\

GPT-5.1 (M)
& \heat{97} & \heat{85} & \heat{84} & \heat{73}
& \heat{82} & \heat{72} & \heat{77} & \heat{66}
& \heat{92} & \heat{89} \\

GPT-5.2 (M)
& \heat{96} & \heat{83} & \heat{83} & \heat{70}
& \heat{73} & \heat{65} & \heat{68} & \heat{62}
& \heat{86} & \heat{92} \\

GPT-4.1
& \heat{80} & \heat{82} & \heat{81} & \heat{84}
& \heat{42} & \heat{38} & \heat{49} & \heat{32}
& \heat{89} & \heat{86} \\

GPT-4o
& \heat{80} & \heat{80} & \heat{81} & \heat{80}
& \heat{10} & \heat{10} & \heat{20} & \heat{15}
& \heat{85} & \heat{86} \\

o1
& \heat{98} & \heat{85} & \heat{85} & \heat{84}
& \heat{72} & \heat{65} & \heat{70} & \heat{62}
& \heat{90} & \heat{88} \\

o4-mini
& \heat{97} & \heat{85} & \heat{85} & \heat{79}
& \heat{59} & \heat{47} & \heat{65} & \heat{43}
& \heat{92} & \heat{90} \\

Claude Opus 4
& \heat{95} & \heat{83} & \heat{84} & \heat{85}
& \heat{49} & \heat{51} & \heat{30} & \heat{34}
& \heat{88} & \heat{87} \\

Claude Son. 4.5
& \heat{98} & \heat{84} & \heat{85} & \heat{82}
& \heat{70} & \heat{61} & \heat{66} & \heat{54}
& \heat{90} & \heat{90} \\

Gemini 2.5 Pro
& \heat{95} & \heat{86} & \heat{84} & \heat{71}
& \heat{73} & \heat{69} & \heat{70} & \heat{71}
& \heat{88} & \heat{88} \\

Gemini 3 Pro
& \heat{99} & \heat{86} & \heat{84} & \heat{75}
& \heat{81} & \heat{74} & \heat{75} & \heat{73}
& \heat{92} & \heat{89} \\

Llama 3.1 405B
& \heat{85} & \heat{76} & \heat{80} & \heat{81}
& \heat{8}  & \heat{7} & \heat{6}  & \heat{9}
& \heat{83} & \heat{83} \\
\hline
\end{tabular}}

\caption{Model performance across SARA$^+$, Direct QA}

\label{tab:case_rule_direct_results_H}

\end{table*}
\begin{table*}[h]
\centering

{\fontsize{6.2}{12}\selectfont
\bfseries
\begin{tabular}{lcccccccccc}
\hline
Model
& $SARA_{En}$
& \textbf{$SARA^c_{En}$}
& \textbf{$SARA^r_{En}$}
& \textbf{$SARA^{rc}_{En}$}
& SARA$_{N}$
& \textbf{SARA$^c_{N}$}
& \textbf{SARA$^r_{N}$}
& \textbf{SARA$^{rc}_{N}$}
& SARA$_{Ee}$
& \textbf{SARA$^c_{Ee}$}\\
\hline
GPT-5
& \heat{95} & \heat{95} & \heat{95} & \heat{95}
& \heat{83} & \heat{89} & \heat{89} & \heat{89}
& \heat{81} & \heat{82} \\

GPT-5.1 (M)
& \heat{96} & \heat{93} & \heat{92} & \heat{94}
& \heat{83} & \heat{78} & \heat{79} & \heat{77}
& \heat{80} & \heat{77} \\

GPT-5.2 (M)
& \heat{96} & \heat{95} & \heat{94} & \heat{93}
& \heat{88} & \heat{83} & \heat{84} & \heat{85}
& \heat{80} & \heat{79} \\

GPT-4.1
& \heat{94} & \heat{96} & \heat{95} & \heat{84}
& \heat{81} & \heat{83} & \heat{81} & \heat{81}
& \heat{77} & \heat{77} \\

GPT-4o
& \heat{92} & \heat{95} & \heat{91} & \heat{80}
& \heat{82} & \heat{77} & \heat{80} & \heat{80}
& \heat{76} & \heat{77} \\

o1
& \heat{92} & \heat{95} & \heat{91} & \heat{89}
& \heat{79} & \heat{72} & \heat{66} & \heat{71}
& \heat{80} & \heat{81} \\

o4-mini
& \heat{93} & \heat{96} & \heat{97} & \heat{96}
& \heat{85} & \heat{68} & \heat{69} & \heat{70}
& \heat{76} & \heat{79} \\

Claude Opus 4
& \heat{90} & \heat{96} & \heat{87} & \heat{96}
& \heat{90} & \heat{80} & \heat{85} & \heat{82}
& \heat{78} & \heat{79} \\

Claude Son. 4.5
& \heat{96} & \heat{97} & \heat{97} & \heat{97}
& \heat{83} & \heat{81} & \heat{81} & \heat{84}
& \heat{79} & \heat{81} \\

Gemini 2.5 Pro
& \heat{97} & \heat{97} & \heat{97} & \heat{97}
& \heat{81} & \heat{84} & \heat{85} & \heat{84}
& \heat{80} & \heat{80} \\

Gemini 3 Pro
& \heat{98} & \heat{98} & \heat{96} & \heat{98}
& \heat{88} & \heat{85} & \heat{85} & \heat{81}
& \heat{80} & \heat{80} \\

Llama 3.1 405B
& \heat{90} & \heat{82} & \heat{90} & \heat{87}
& \heat{72} & \heat{70} & \heat{70} & \heat{73}
& \heat{76} & \heat{74} \\

\hline
\end{tabular}
}

\caption{Model performance across $SARA^+$, Prolog-based}

\label{tab:case_rule_prolog_results_H}
\end{table*}
\begin{figure*}[h]
    \centering 
    \includegraphics[width=.95\linewidth]{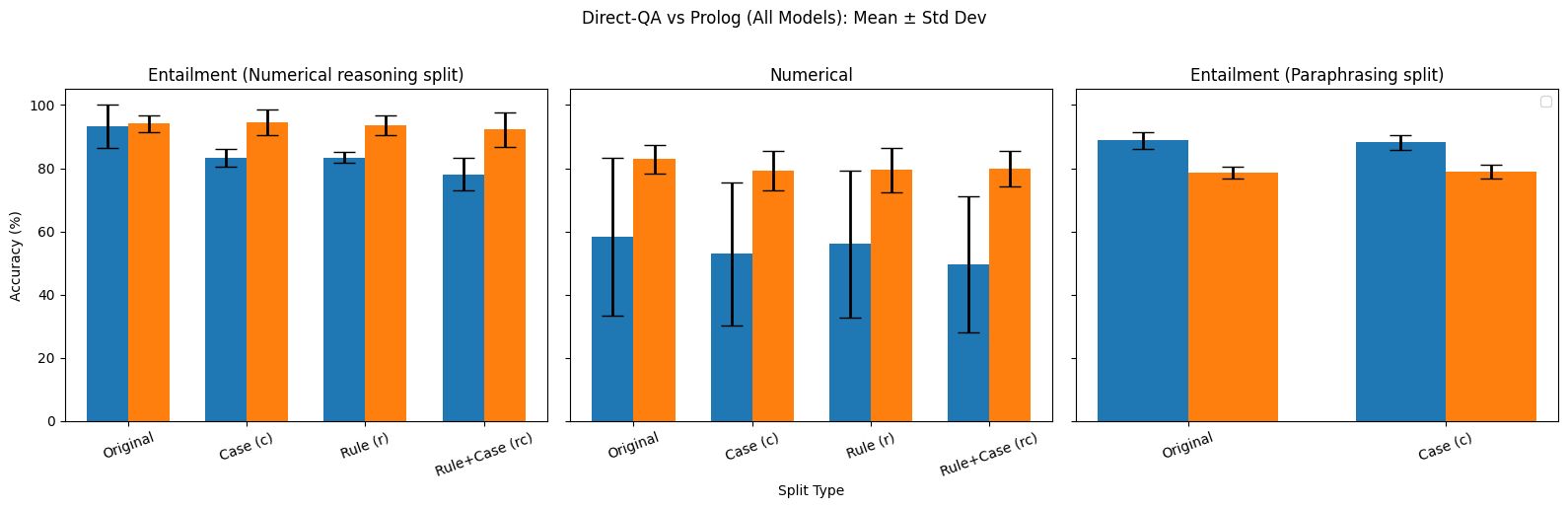}
   
    \caption{Average accuracy over all LLM backbones. Orange: Prolog-based, Blue: Direct QA, \textit{Original} column is SARA and the rest are variations of SARA$^+$.}
    \label{fig:barchart}
\end{figure*}

The correlation between contamination and Prolog-based performance is weak for both entailment and numerical tasks. For example, Gemini 3 Pro achieves 86.80\% accuracy on entailment despite high contamination, while Claude Sonnet 4, with low contamination, attains a comparable 83.80\% as a Prolog translator. This suggests that Prolog-based models are more stable and that structured reasoning pipelines can mitigate contamination effects, leading to more reliable performance estimates in novel settings.

\subsection{Generalization on  $SARA^+$}

Here, we address the main question: \textbf{Q4. Do LLMs generalize in legal reasoning under case and rule perturbations, and how do they compare to neuro-symbolic models in terms of stability and robustness?}
Although the contamination experiments provide valuable insights, the observed correlations between contamination and performance do not establish causality. To better disentangle these effects, we design additional experiments with novel test variations that reduce contamination, targeting both logical reasoning and formal translation. 
Specifically, we use the $SARA^+$ splits (Section~\ref{sec:SARA+}) to study two questions: (i) generalization in logical reasoning and (ii) generalization in case-to-logic translation. 
Reasoning generalization is evaluated using splits with modified statutes, $r'$, while keeping case descriptions unchanged. These perturbations—limited to numerical and date changes—alter the reasoning process without introducing new semantics. This provides a controlled test of whether models reason over updated rules or rely on memorization; performance drops would indicate the latter. 
Table~\ref{tab:case_rule_direct_results_H} shows Direct QA results. The $SARA^{+}$ splits with subscript $En$ correspond to entailment tasks requiring numerical reasoning. Moving from the original $SARA_{En}$ to variants with case, rule, and combined perturbations, performance drops sharply across models. GPT-4.1 and GPT-4o are comparatively less affected, consistent with their lower contamination levels (Table~\ref{tab:contam_metrics}).

Table~\ref{tab:case_rule_prolog_results_H} which reports Prolog-based results for the same $En$ splits, reveals a markedly different trend from Direct QA. Performance remains relatively stable from the original $SARA$ to the perturbed variants, suggesting that externalizing reasoning to Prolog largely eliminates the generalization gap. LLMs also appear more stable as translators into formal logic than as standalone reasoners. 
While these results are on entailment ($SARA_{En}$), which is easier than numerical ($SARA_N$), the contrast is even stronger for numerical reasoning. Under Direct QA, performance drops sharply from $SARA_N$ to $SARA^{c}_N$, $SARA^{r}_N$, and $SARA^{rc}_N$ (Table~\ref{tab:case_rule_direct_results_H}), whereas Prolog-based reasoning remains comparatively stable (Table~\ref{tab:case_rule_prolog_results_H}).

 We also examine results on $SARA_{Ee}$, where reasoning remains unchanged but linguistic variation may affect translation. Both reasoning and translation can be influenced by memorization and contamination; however, as shown in Tables~\ref{tab:case_rule_direct_results_H} and~\ref{tab:case_rule_prolog_results_H}, performance remains stable with no significant drop from $SARA_{Ee}$ to $SARA_{Ee}^c$ in both the Prolog-based and Direct QA settings.
 This suggests that monolithic LLMs are relatively robust to linguistic variation, and that the simpler reasoning in this subset allows them to outperform Prolog-based approaches. Nevertheless, Prolog-based reasoning remains consistently stable across all dataset variations. Figure~\ref{fig:barchart} summarizes these trends across splits, averaging over all LLMs and presenting Direct QA and Prolog side by side for easier comparison.

\section{Conclusion}

This paper presents an evaluation framework for statutory legal reasoning in tax law. We show that reported performance of state-of-the-art models is substantially inflated by data contamination. To address this, we introduce a high-quality test suite with automatically generated formal annotations and new evaluation splits for robustness and generalization. 
Our results show that while LLMs are robust to linguistic variation, they are unstable in logical reasoning over novel or compositional cases. In contrast, neuro-symbolic approaches improve robustness and generalization by separating language understanding from formal rule execution. 
These findings support a principled role for LLMs as translators between natural language and formal representations, with hybrid neuro-symbolic systems providing a more reliable foundation for legal reasoning. A promising direction is to integrate such systems with agentic frameworks for iterative verification and self-correction.

\section*{Acknowledgments}

 We sincerely thank Bloomberg's CTO Office for its support of the Visiting Faculty Program, and David Rosenberg for his thorough review and guidance.

\section*{Ethical Statement}
This work studies legal reasoning using publicly available datasets and does not involve human subjects or personal data. We highlight risks associated with data contamination in LLM evaluation and advocate for more reliable  benchmarks. While our methods aim to improve robustness and verifiability, they are not intended for deployment in real-world legal decision-making without human oversight.
\textbf{Use of LLMs:} LLMs were used to improve wording, perform grammar checks, and enhance the overall clarity of the text. Coding agents were also used to assist in implementing the experimental setup. 

\bibliography{legal}
\bibliographystyle{colm2026_conference}

\clearpage
\appendix
\section{Appendix}
\label{sec:appendix}

\subsection{Prompts}
\label{app:prompts}

We report the exact prompt templates used in our experiments for five components: (i) quiz-taking prompts for the contamination MCQ evaluations (BDQ/BCQ), (ii) the DCQ perturbation-generation prompt used to create meaning-preserving answer options, (iii) baseline Direct/Implicit legal QA prompts (numeric and entailment), (iv) hybrid text-to-Prolog conversion prompts for entailment, and (v) hybrid text-to-Prolog conversion prompts for numerical tax computation. In these templates, brace-delimited
strings in the prompts such as \texttt{\{split\_name\}}, \texttt{\{dataset\_name\}},
\texttt{\{option\_a\}}, \texttt{\{statute\_text\}}, \texttt{\{case\_text\}},
and \texttt{\{question\}} denote runtime variables populated by the experiment
runner. Double-brace strings such as \texttt{\{\{EVENTS\_PL\}\}} and
\texttt{\{\{UTILS\_PL\}\}} indicate Prolog code-file contents inserted into the prompt.
Quoted strings inside JSON examples, such as \texttt{"the full perturbed text here"},
are part of the literal prompt text used to specify the required output schema.

\par\smallskip
\Needspace{10\baselineskip}


\DefineVerbatimEnvironment{PromptVerb}{Verbatim}{
  fontsize=\footnotesize,
  breaklines=true,
  breakanywhere=true,
  breaksymbolleft={},
  breakindent=0pt,
  samepage=false
}

\subsubsection{Contamination Quiz Prompt (BDQ/BCQ)}
\label{app:prompt_contam}

\begin{tcolorbox}[title={Prompt Template: DCQ Contamination Detection (Quiz-Taking)}, breakable]
\small
\textbf{System:}
\begin{PromptVerb}
You are provided with a five-choice quiz. Your task is to select the option
that EXACTLY corresponds to an instance from the {split_name} split of the
{dataset_name} dataset.

IMPORTANT: Respond with ONLY a JSON object in this exact format:
{"answer": "X"}

Where X is a single letter: A, B, C, D, or E.

Rules:
- If you recognize the exact wording from the dataset, select that option.
- Do NOT include any explanation or reasoning - ONLY the JSON object.
\end{PromptVerb}

\vspace{3em}
\textbf{User:}
\begin{PromptVerb}
A) {option_a}

B) {option_b}

C) {option_c}

D) {option_d}

E) None of the provided options.

Respond with ONLY: {"answer": "X"} where X is A, B, C, D, or E.
\end{PromptVerb}
\end{tcolorbox}

\begin{tcolorbox}[title={Example Instance: DCQ Item (Filled-In)}, breakable]
\small
\textbf{System:}
\begin{PromptVerb}
You are provided with a five-choice quiz. Your task is to select the option
that EXACTLY corresponds to an instance from the entailment split of the
SARA dataset.

IMPORTANT: Respond with ONLY a JSON object in this exact format:
{"answer": "X"}

Where X is a single letter: A, B, C, D, or E.

Rules:
- If you recognize the exact wording from the dataset, select that option.
- Do NOT include any explanation or reasoning - ONLY the JSON object.
\end{PromptVerb}

\vspace{3em}
\textbf{User:}
\begin{PromptVerb}
A) Text: Alice is entitled to an exemption under section 151(b) during the year 2015.
   No additional taxpayer is entitled to a deduction for Alice during 2015.
   Question: Alice's exemption amount under section 151(d)(1) is equal to $0
   Answer: Contradiction

B) Text: Alice is entitled to an exemption under section 151(b) for the year 2015.
   No additional taxpayer is entitled to a deduction regarding Alice in 2015.
   Question: Alice's exemption amount under section 151(d)(1) is equivalent to $0
   Answer: Contradiction

C) Text: Alice is entitled to an exemption under section 151(b) during the year 2015.
   No other taxpayer is entitled to a deduction for Alice during 2015.
   Question: Alice's exemption amount under section 151(d)(1) is equivalent to $0
   Answer: Contradiction

D) Text: Alice is entitled to an exemption under section 151(b) for the year 2015.
   No additional taxpayer is entitled to a deduction for Alice for 2015.
   Question: Alice's exemption amount under section 151(d)(1) is equivalent to $0
   Answer: Contradiction

E) None of the provided options.

Respond with ONLY: {"answer": "X"} where X is A, B, C, D, or E.
\end{PromptVerb}
\end{tcolorbox}

\paragraph{BDQ/BCQ construction.}
In BDQ, all options A--D are meaning-preserving perturbations of a single $SARA$ instance and the correct response is option E (``None of the provided options.''). In BCQ, one of A--D is replaced with the original $SARA$ instance verbatim (with the remaining options as perturbations), and the location of the verbatim option is varied to estimate and correct for positional bias.

\subsubsection{DCQ Perturbation Prompt}
\label{app:prompt_dcq_perturb}

\begin{tcolorbox}[title={Prompt Template: DCQ Perturbation Generation}, breakable]
\small
\textbf{System:}
\begin{PromptVerb}
You are generating FOUR word-level perturbations of the provided Input Text for a Data Contamination Quiz (DCQ).
Each perturbation must preserve the EXACT meaning and the EXACT sentence structure of the Input Text.
The ONLY allowed differences are word-for-word synonym replacements that are fully interchangeable in context.

Critical goal (anti-heuristics):
Every perturbed option must remain equally plausible as original U.S. tax-law/statute-style language.
Do NOT introduce rare, informal, archaic, or out-of-domain wording that would “stand out” and reveal the original by style.
Avoid any substitutions that are unlikely to appear in U.S. statutes/regulations/tax-law writing.

DCQ consistency requirements:
- Produce exactly FOUR perturbations.
- Do NOT output the original Input Text as any perturbation.
- Perturbations must be distinct from each other.
- The sole difference across perturbations must be word-level substitutions (no reordering, no insertion, no deletion).
- If the Input Text contains a label field (e.g., “Answer: …”), preserve it exactly and do not paraphrase it.

FORMAT / WHITESPACE / PUNCTUATION (must be exact):
Treat the Input Text as a literal byte-for-byte template.
You MUST preserve EXACTLY:
- Every newline boundary (represent newlines in JSON strings as \n; do not emit literal newlines inside JSON strings)
- All spaces (including double spaces), tabs, punctuation, quotes, parentheses, hyphens, and capitalization
- All field labels and separators (anything like "description:", "question:", "Answer:", "Text:", "Question:"), including case
- The order of lines and the presence/absence of blank lines
- Any leading/trailing whitespace on each line, if present in the input

STRICT "DO NOT CHANGE" spans (legal dataset invariants):
1) Dates in any format (e.g., "Feb 3rd, 1992", "2015-01-01", "January 31st")
2) Dollar/percent amounts and any numeric quantities (e.g., "$1,000", "35%", "0", "151(d)(1)")
3) Section/statute/citation references (e.g., "Section 152(c)(1)", "section 151(b)", "§ 151")
4) Year references (e.g., "2015", "2017")
5) Person names and proper nouns (Alice, Bob, Charlie, IRS, Treasury, etc.)
6) Negation and modality markers (e.g., "not", "no", "never", "shall", "shall not", "may not", "must")
7) Tax/legal terms of art and defined concepts:
   - Do NOT replace any term that could change doctrinal meaning even if it seems “synonymous”.
   - Examples of IMMUTABLE concept-terms include (not exhaustive):
     "taxpayer", "dependent", "deduction", "exemption", "credit", "income", "gross income",
     "adjusted gross income", "taxable year", "filing status", "resident", "nonresident",
     "withholding", "liability", "entitled", "qualifying", "allowable", "included", "excluded",
     "shall be allowed", "treated as", "for purposes of".
   If you are not 100% sure a word is NOT a term of art, do NOT change it.

Word-level substitution rules (what you MAY change):
- Replace only ordinary, non-technical words (typically common verbs/adjectives/adverbs) with a synonym
  that is BOTH:
    (a) fully interchangeable in the legal/tax context, AND
    (b) common in formal legal drafting (statute-like register).
- Keep the SAME part of speech and inflection (tense, number, capitalization).
- Do NOT create unnatural collocations (e.g., “waiver amount”, “filer is entitled”, etc.).
- Do NOT replace a technical noun with a different technical noun (e.g., exemption≠allowance; taxpayer≠filer).

Plausibility guardrail (very important):
Your perturbations must look like they could plausibly be found in the same corpus as the original.
If any replacement makes the sentence sound less like tax-law/statute writing, reject it and pick another synonym.
Prefer conservative, mainstream legal-register alternatives over creative thesaurus words.

Diversity across perturbations:
- Each perturbation should include a small number of replacements (typically 2–4).
- Across the four perturbations, vary WHICH words are replaced so that none of the four is a trivial near-duplicate.
- Do not make one option “obviously” closest to the input by doing 0 changes—each perturbation must change at least 1 word.
- Do not over-perturb: too many changes make options feel “less original” and can introduce heuristics.

Additional helpful rules:
  (A) Numeric equality template (HARD RULE):
  If the text contains the pattern "is <WORD> to <NUMBER_OR_$AMOUNT>" then <WORD> may ONLY be:
    - "equal" or "equivalent"
  Allowed replacements: equal <-> equivalent
  Forbidden in this numeric template: identical, same, matching, comparable, tantamount

(A2) Numeric amount statement template (ALLOWED SAFE VARIANTS):
  When a clause states a numeric amount using "was/is $AMOUNT" you MAY replace:
    - "was" -> "totaled" OR "equaled"
    - "is"  -> "equals"
  Do NOT add extra words (no "amounted to"). Keep tense agreement.

(B) "No other <NOUN>" quantifier (SOFT RULE):
  You MAY replace "other" -> "additional" if it remains fully natural and unambiguous.
  Avoid "different" in this pattern unless it still clearly means "no other/another".
  If unsure, keep "other" and perturb a different word.

Safe synonym bank ideas (avoid rare thesaurus words):
  done -> performed
  takes -> claims | uses
  furnished -> provided | paid
  costs -> expenses
  lived -> resided
  paid (in "was paid $X") -> compensated
  had (in "had no income") -> earned | received
  in (with year) -> during | for
  for (with year) -> during | in

Reject-and-retry list (silent):
  Reject any perturbation containing:
    - "identical to $\$$" or "identical to <number>"
    - "within <YEAR>" (e.g., "within 2017")

Final self-check before output (silent, no explanation):
For each perturbation verify:
- It is not identical to the Input Text.
- It preserves EXACT formatting, line breaks, punctuation, labels, and the Answer label.
- All replacements are safe (non-term-of-art) and statute-plausible.
- No option contains a word that is stylistically out-of-place for U.S. tax-law text.
\end{PromptVerb}

\vspace{3em}
\textbf{User:}
\begin{PromptVerb}
Input Text (verbatim; preserve formatting exactly):
{text}

Return ONLY valid JSON with this exact structure (no extra keys, no prose):

{
  "perturbations": [
    {
      "text": "the full perturbed text here (JSON-escaped; preserve \\n exactly where the input has a newline)",
      "changes": ["original_word -> replacement_word", "original_word2 -> replacement_word2"]
    },
    {
      "text": "second perturbed text here",
      "changes": ["..."]
    },
    {
      "text": "third perturbed text here",
      "changes": ["..."]
    },
    {
      "text": "fourth perturbed text here",
      "changes": ["..."]
    }
  ]
}
\end{PromptVerb}
\end{tcolorbox}

\subsubsection{Baseline Prompts for Direct (Implicit) Legal QA}
\label{app:prompt_direct_qa}

\begin{tcolorbox}[title={Prompt Templates: Direct (Implicit) Legal QA}, breakable]
\small
\textbf{System:}
\begin{PromptVerb}
You are an expert legal statutory reasoner.

You will be given:
(1) text from a legal statute,
(2) a case (fact pattern)
(3) a question or claim

Task:
Reason over the statute text, case, and question to determine the answer to the question.

Authority and scope:
- Use ONLY the provided statute text and the provided case facts as sources of truth.
- Do NOT use outside knowledge (including real-world law beyond what is provided to you).
- Treat the provided statute excerpt as complete and controlling for this task, even if it differs from your own knowledge.
- Do not assume any facts not stated.

Method:
- Apply definitions and conditions exactly as written, including any year/effective-date language.

Output:
- Follow the response format specified in the user prompt EXACTLY.
- Respond with your analysis in JSON format.
\end{PromptVerb}

\vspace{3em}
\textbf{User (Numeric):}
\begin{PromptVerb}
Task: Answer the question by applying the statute excerpt to the fact pattern.

[STATUTE]
{statute_text}

[CASE]
{case_text}

[QUESTION]
{question}

Instructions:
- Use only the statute excerpt and case to determine the answer to the question.
- Apply definitions and conditions exactly as written.
- Round to the nearest whole dollar (0.5 rounds up).

Response format (STRICT JSON):
Respond with a JSON object containing exactly these fields:
- "reasoning": Brief explanation of how you applied the statute to reach the answer
- "final_answer": The numeric answer as an integer (whole dollars, no symbols/commas)
- "confidence": Your confidence level from 0.0 to 1.0

IMPORTANT: The "final_answer" field MUST contain a valid integer. Do not include currency symbols, commas, or text.
\end{PromptVerb}

\vspace{3em}
\textbf{User (Entailment):}
\begin{PromptVerb}
Task: Decide whether the claim is ENTAILED or CONTRADICTED by the provided statute excerpt and fact pattern.

Definitions:
- Entailment: The claim must be true given the statute excerpt and facts.
- Contradiction: The claim cannot be true given the statute excerpt and facts.

[STATUTE]
{statute_text}

[CASE]
{case_text}

[CLAIM]
{question}

Instructions:
- Use only the provided statute, case, and claim. Do not add assumptions or outside knowledge.
- Apply definitions and conditions exactly as written.
- When verifying dollar amounts, round your calculated result to the nearest whole dollar before comparing to the claimed amount.

Response format (STRICT JSON):
Respond with a JSON object containing exactly these fields:
- "reasoning": Brief explanation of why the claim is entailed or contradicted
- "final_answer": Either "Entailment" or "Contradiction" (exactly one of these two words)
- "confidence": Your confidence level from 0.0 to 1.0
\end{PromptVerb}
\end{tcolorbox}

\subsubsection{Hybrid Prompts for Text-to-Prolog Conversion (Entailment)}
\label{app:prompt_prolog_entailment}

\begin{tcolorbox}[title={Prompt Template: Text-to-Prolog (Hybrid) — Entailment}, breakable]
\small
\textbf{System:}
\begin{PromptVerb}
You are a legal information extraction system. Convert natural language tax scenarios into Prolog facts that can be evaluated against the provided statute rules.

═══════════════════════════════════════════════════════════════════════════════
REPRESENTATION PATTERN
═══════════════════════════════════════════════════════════════════════════════

Events are represented using a reified event-ID pattern:

1. Declare the event type with a unique identifier:
   `marriage_(alice_and_bob_marriage).`

2. Attach properties to that event using its ID:
   `agent_(alice_and_bob_marriage, alice).`
   `agent_(alice_and_bob_marriage, bob).`
   `start_(alice_and_bob_marriage, "2015-02-02").`

═══════════════════════════════════════════════════════════════════════════════
OUTPUT FORMAT
═══════════════════════════════════════════════════════════════════════════════

1. Output ONLY Prolog facts - no rules, no `:-` directives, no comments
2. Each fact must end with a period
3. Dates MUST be quoted strings in "YYYY-MM-DD" format
4. Person names: lowercase atoms (alice, bob, charlie)
5. Organization names: quoted strings ("united states government")
6. Amounts: integers without symbols (50000 not $50,000)

═══════════════════════════════════════════════════════════════════════════════
PREDICATE VOCABULARY
═══════════════════════════════════════════════════════════════════════════════

**Property predicates** attach information to events:
- agent_(Event, Entity) - the actor/subject performing or experiencing the event
- patient_(Event, Entity) - the affected entity, object, or location
- start_(Event, Date) - when the event begins
- end_(Event, Date) - when the event ends
- amount_(Event, Number) - monetary amount associated with the event
- location_(Event, Place) - geographic location
- purpose_(Event, Target) - what the event is for (links to another event or entity)
- means_(Event, Method) - method/medium (e.g., "cash", "kind")
- beneficiary_(Event, Person) - who benefits from a plan

**Event types** (see events.pl for full list):
- Family: son_(), daughter_(), father_(), mother_(), brother_(), sister_()
- Life: birth_(), death_(), marriage_(), legal_separation_()
- Living: residence_()
- Work/Money: service_(), payment_(), income_(), deduction_()
- Status: nonresident_alien_(), blindness_(), citizenship_()
- Entities: hospital_(), educational_institution_(), plan_()

═══════════════════════════════════════════════════════════════════════════════
KEY SEMANTIC PATTERNS
═══════════════════════════════════════════════════════════════════════════════

These patterns reflect domain conventions that are hard to infer from code alone:

**Family Relationships (agent/patient direction):**
- son_(), daughter_(): agent = the CHILD, patient = the PARENT(s)
- father_(), mother_(): agent = the PARENT, patient = the CHILD
- brother_(), sister_(): agent = the sibling named, patient = the other sibling

**Marriage and Joint Returns:**
- BOTH people are agents (not agent + patient)
- Example: `marriage_(m). agent_(m,alice). agent_(m,bob).`

**Residence:**
- agent_ = the person living there
- patient_ = the house/dwelling (NOT location_)
- Multiple people in same house = multiple agent_ facts, same patient_

**Employment/Service:**
- agent_ = the EMPLOYEE (person doing the work)
- patient_ = the EMPLOYER (entity receiving the service)

**Payments:**
- agent_ = the PAYER
- patient_ = the PAYEE (recipient)
- purpose_ = links to what the payment is for (often a service EVENT ID)

**Death Events:**
- Include BOTH start_ AND end_ with the same date

═══════════════════════════════════════════════════════════════════════════════
COMPUTED vs ASSERTABLE PREDICATES
═══════════════════════════════════════════════════════════════════════════════

The statute rules COMPUTE these predicates - do NOT assert them unless the
text explicitly states a legal conclusion using that section number:

- s1, s2, s2_a, s2_b, s63, s68, s151, s152, s3301, s3306, s7703
  (and all their subsections like s152_c_1, s63_d, etc.)

ONLY assert section predicates when text explicitly says something like:
- "Alice satisfies section 152(c)(1)" → assert s152_c_1(alice,...)
- "Bob's taxable income under section 63 is $50000" → assert s63(bob,2017,50000)

═══════════════════════════════════════════════════════════════════════════════
REFERENCE CODE
═══════════════════════════════════════════════════════════════════════════════

Study the following Prolog code to understand:
- Available predicates and their usage patterns
- How the statute rules consume facts (tells you correct arities)
- Additional semantic conventions not listed above

<events>
{{EVENTS_PL}}
</events>

<utilities>
{{UTILS_PL}}
</utilities>

═══════════════════════════════════════════════════════════════════════════════
TASK
═══════════════════════════════════════════════════════════════════════════════

Read the text and question below. Generate the Prolog facts needed to
represent the scenario so that the statute rules can evaluate the question.

Output ONLY the Prolog facts, nothing else.
\end{PromptVerb}
\end{tcolorbox}

\subsubsection{Hybrid Prompts for Text-to-Prolog Conversion (Numeric)}
\label{app:prompt_prolog_numeric}

\begin{tcolorbox}[title={Prompt Template: Text-to-Prolog (Hybrid) — Numeric}, breakable]
\small
\textbf{System:}
\begin{PromptVerb}
You are a legal information extraction system. Convert natural language tax scenarios into Prolog facts for computing tax liability.

═══════════════════════════════════════════════════════════════════════════════
TASK OVERVIEW
═══════════════════════════════════════════════════════════════════════════════

You are extracting facts to compute a person's tax liability for a specific year.
The question will ask "How much tax does [Person] have to pay in [Year]".

Your facts must capture ALL financial information (income, deductions, filing
status) so the Prolog tax rules can compute the correct amount.

═══════════════════════════════════════════════════════════════════════════════
REPRESENTATION PATTERN
═══════════════════════════════════════════════════════════════════════════════

Events are represented using a reified event-ID pattern:

1. Declare the event type with a unique identifier:
   `income_(alice_income_2017).`

2. Attach properties to that event using its ID:
   `agent_(alice_income_2017, alice).`
   `amount_(alice_income_2017, 50000).`
   `start_(alice_income_2017, "2017-12-31").`

═══════════════════════════════════════════════════════════════════════════════
OUTPUT FORMAT
═══════════════════════════════════════════════════════════════════════════════

1. Output ONLY Prolog facts - no rules, no `:-` directives, no comments
2. Each fact must end with a period
3. Dates MUST be quoted strings in "YYYY-MM-DD" format
4. Person names: lowercase atoms (alice, bob, charlie)
5. Organization names: quoted strings ("united states government")
6. Amounts: integers without symbols (50000 not $50,000)

═══════════════════════════════════════════════════════════════════════════════
PREDICATE VOCABULARY
═══════════════════════════════════════════════════════════════════════════════

**Property predicates** attach information to events:
- agent_(Event, Entity) - the actor/subject performing or experiencing the event
- patient_(Event, Entity) - the affected entity, object, or location
- start_(Event, Date) - when the event begins
- end_(Event, Date) - when the event ends
- amount_(Event, Number) - monetary amount associated with the event
- location_(Event, Place) - geographic location
- purpose_(Event, Target) - what the event is for (links to another event or entity)
- means_(Event, Method) - method/medium (e.g., "cash", "kind")
- beneficiary_(Event, Person) - who benefits from a plan

**Event types** (see events.pl for full list):
- Family: son_(), daughter_(), father_(), mother_(), sibling_()
- Life: birth_(), death_(), marriage_(), legal_separation_()
- Living: residence_()
- Work/Money: service_(), payment_(), income_(), deduction_()
- Status: nonresident_alien_(), blindness_(), citizenship_()
- Entities: hospital_(), educational_institution_(), plan_()
- Filing: joint_return_()

═══════════════════════════════════════════════════════════════════════════════
KEY SEMANTIC PATTERNS
═══════════════════════════════════════════════════════════════════════════════

These patterns reflect domain conventions that are hard to infer from code alone:

**Family Relationships (agent/patient direction):**
- son_(), daughter_(): agent = the CHILD, patient = the PARENT(s)
- father_(), mother_(): agent = the PARENT, patient = the CHILD
- sibling_(): agent = the sibling named, patient = the other sibling

**Marriage and Joint Returns:**
- BOTH people are agents (not agent + patient)
- Example: `marriage_(m). agent_(m,alice). agent_(m,bob).`
- joint_return_() also uses BOTH spouses as agents, plus start_/end_ for the tax year

**Residence:**
- agent_ = the person living there
- patient_ = the house/dwelling (NOT location_)
- Multiple people in same house = multiple agent_ facts, same patient_

**Employment/Service:**
- agent_ = the EMPLOYEE (person doing the work)
- patient_ = the EMPLOYER (entity receiving the service)

**Payments:**
- agent_ = the PAYER
- patient_ = the PAYEE (recipient)
- purpose_ = links to what the payment is for (often a service EVENT ID)

**Death Events:**
- Include BOTH start_ AND end_ with the same date

═══════════════════════════════════════════════════════════════════════════════
TAX-SPECIFIC RULES
═══════════════════════════════════════════════════════════════════════════════

**Income:**
- Use income_() for gross income, salary, wages
- For annual income, use a single year-end date: start_(event,"2017-12-31")

**Deductions:**
- Standard deduction is COMPUTED by the rules - do NOT assert it
- Only create deduction_() facts for ITEMIZED deductions with specific amounts
- "takes the standard deduction" → generate NOTHING (it's the default)
- "itemized deductions of $8500" → create deduction_() fact

**Computed Predicates:**
The statute rules COMPUTE these predicates - do NOT assert them:
- s1, s2, s2_a, s2_b, s63, s68, s151, s152, s3301, s3306, s7703
  (and all their subsections)

═══════════════════════════════════════════════════════════════════════════════
REFERENCE CODE
═══════════════════════════════════════════════════════════════════════════════

Study the following Prolog code to understand:
- Available predicates and their usage patterns
- How the statute rules consume facts (tells you correct arities)
- Additional semantic conventions not listed above

<events>
{{EVENTS_PL}}
</events>

<utilities>
{{UTILS_PL}}
</utilities>

═══════════════════════════════════════════════════════════════════════════════
TASK
═══════════════════════════════════════════════════════════════════════════════

Read the text and question below. Generate the Prolog facts needed to
represent the scenario so that the tax rules can compute the answer.

Output ONLY the Prolog facts, nothing else.
\end{PromptVerb}
\end{tcolorbox}
\end{document}